\documentclass[10pt,twocolumn,letterpaper]{article}

\usepackage{iccv}
\usepackage{times}
\usepackage{epsfig}
\usepackage{graphicx}
\usepackage{amsmath}
\usepackage{amssymb}
\usepackage{multirow}
\usepackage{bbm}
\usepackage{enumitem}
\usepackage{booktabs}
\usepackage[table,dvipsnames]{xcolor}
\usepackage{array}
\newcolumntype{L}[1]{>{\raggedright\let\newline\\\arraybackslash\hspace{0pt}}p{#1}}

\usepackage[breaklinks=true,bookmarks=false]{hyperref}
\usepackage[capitalize]{cleveref}
\crefname{section}{Sec.}{Secs.}
\Crefname{section}{Section}{Sections}
\Crefname{table}{Table}{Tables}
\crefname{table}{Tab.}{Tabs.}
\newcommand{\smallsec}[1]{\vspace{0.04in} \noindent {\bf #1.}}

\DeclareMathOperator*{\argmax}{arg\,max}
\DeclareMathOperator*{\argmin}{arg\,min}

\newcommand{\MU}{$\textbf{UUU}${}}

\newcommand{\SU}{$\textbf{UU}$}{}

\newcommand{\LU}{$\textbf{U}$}{}

\newcommand{\MF}{$\textbf{FFF}$}{}
\newcommand{\SF}{$\textbf{FF}$}{}
\newcommand{\LF}{$\textbf{F}$}{}

\iccvfinalcopy 

\def\iccvPaperID{3957} 

\ificcvfinal\fi

\begin{document}

\title{UFO: A unified method for controlling Understandability and Faithfulness Objectives in concept-based explanations for CNNs}

\author{Vikram V. Ramaswamy, Sunnie S. Y. Kim, Ruth Fong, Olga Russakovsky\\
Princeton University\\
{\tt\small \{vr23, suhk, ruthfong, olgarus\}@cs.princeton.edu}
}

\maketitle
\ificcvfinal\thispagestyle{empty}\fi

\begin{abstract}
    Concept-based explanations for convolutional neural networks (CNNs) aim to explain model behavior and outputs using a pre-defined set of semantic concepts (e.g., the model recognizes scene class ``bedroom'' based on the presence of concepts ``bed'' and ``pillow'').
    However, they often do not faithfully (i.e., accurately) characterize the model's behavior and can be too complex for people to understand.
    Further, little is known about how faithful and understandable different explanation methods are, and how to control these two properties.
    In this work, we propose UFO, a unified method for controlling Understandability and Faithfulness Objectives in concept-based explanations. UFO formalizes understandability and faithfulness as mathematical objectives and unifies most existing concept-based explanations methods for CNNs. Using UFO, we systematically investigate how explanations change as we turn the knobs of faithfulness and understandability.
    Our experiments demonstrate a faithfulness-vs-understandability tradeoff: increasing understandability reduces faithfulness.
    We also provide insights into the ``disagreement problem'' in explainable machine learning, by analyzing when and how concept-based explanations disagree with each other.
\end{abstract}

\section{Introduction}
\label{sec:intro}

As convolutional neural networks (CNNs) start to be used to make consequential decisions, such as those in medical diagnosis and treatment, there is a need for these models to be interpretable. Over the past decade, many model explanation methods have been proposed, ranging from local methods that explain a single prediction, e.g., by highlighting relevant pixels in an input image~\cite{chattopadhay2018gradcamplusplus,fong19understanding,Petsiuk2018rise,selvaraju2017gradcam,simonyan2013saliency,zeiler2014visualizing,zhang2016excitation,zhou2016cam}, to global methods that provide a higher-level understanding of what the model has learned and how it predicts a certain target class~\cite{bau2017netdissect,fong2018net2vec,ramaswamy2022elude,zhou2018ibd}. However, explanations often \emph{disagree} with each other~\cite{krishna2022disagreement},
making it difficult for users to choose which explanation method to use.

\begin{figure}[t]
    \centering
    \includegraphics[width=\linewidth]{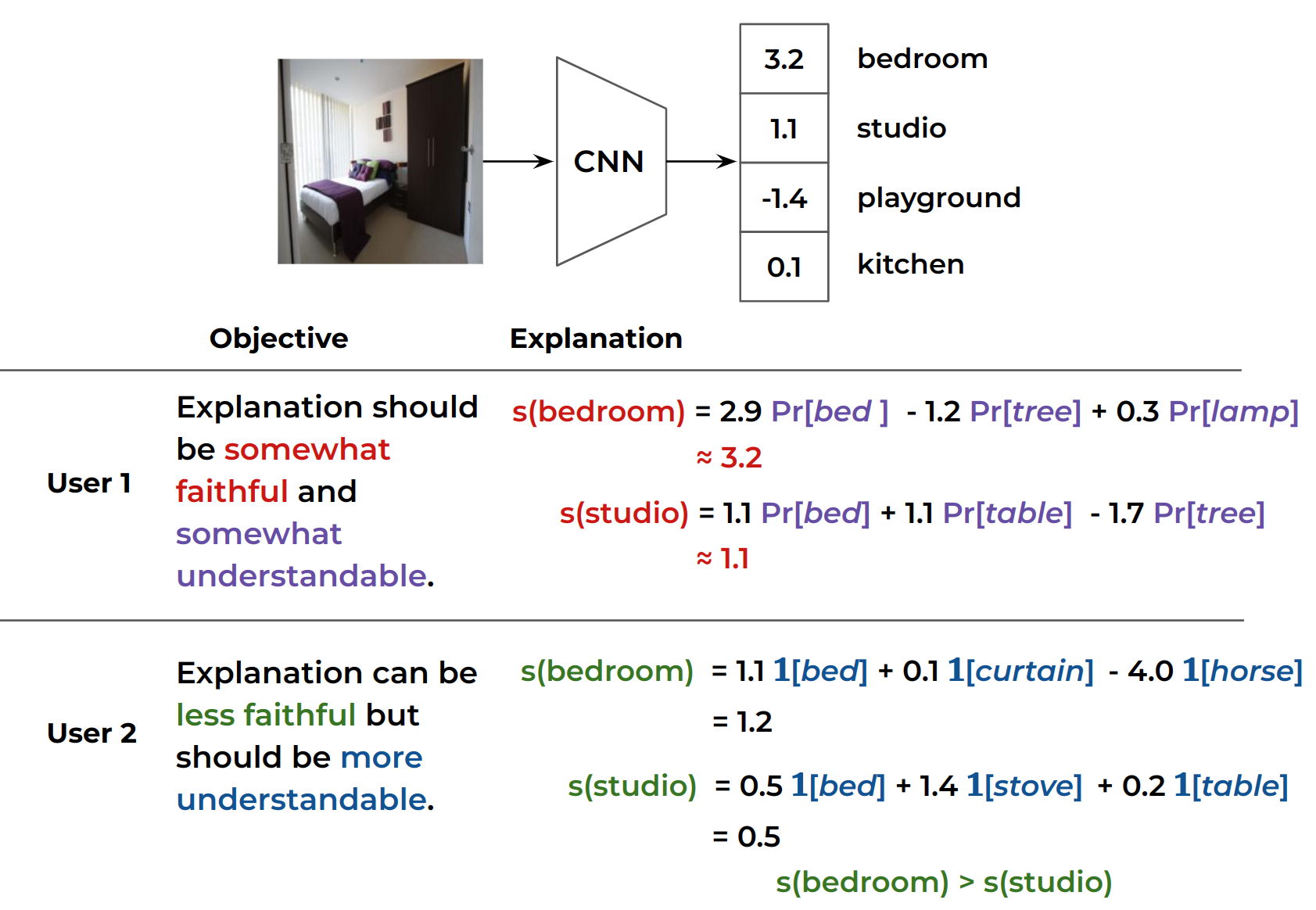}
    \caption{Consider a CNN that outputs probabilities for the target classes (e.g., bedroom, studio). 
    For explanations, different users might have different objectives of (1) ``faithfulness'': matching probabilities of target classes (user 1, \textcolor{Maroon}{``somewhat faithful''}) vs. just matching the output class (user 2, \textcolor{ForestGreen}{``less faithful''}) and (2) ``understandability'': encoding concepts as probabilities (user 1, \textcolor{Purple}{``somewhat understandable''}) or binary values (user 2, \textcolor{NavyBlue}{``more understandable''}).
    Our proposed method unifies existing concept-based explanation methods for CNNs and enables users to control ``faithfulness'' and ``understandability.''}
    \label{fig:pull}
\end{figure}

One reason for this disagreement is the lack of consensus among researchers about the goals of these methods. While most methods attempt to optimize objectives like ``faithfulness'' and ``understandability,'' these terms are not well defined and can be translated into different mathematical objectives. For example, ``faithfulness'' requires that the explanation accurately describes the model's behaviour, but this can be thought of at different levels: Do we care about the distribution of all class scores from the model (as in \cite{zhou2018ibd,koh2020conceptbottleneck})? Or just the score for the target class (as in \cite{ramaswamy2022elude})? 
Similarly, ``understandability'' can mean different things to different explanation users: 
AI experts (e.g., model developers) may find an explanation that encodes concepts as probabilities understandable, while non-experts (e.g., lay end-users) may not and thus, prefer an explanation that encodes concepts as binary values~\cite{kim2023helpmehelptheai}.
In \cref{fig:pull}, we show how different users might want different levels of faithfulness and understandability in model explanations. 

In this work, we propose UFO, a novel concept-based CNN explanation method that formalizes (U)nderstandability and (F)aithfulness as mathematical (O)bjectives. 
UFO unifies existing methods~\cite{bau2017netdissect,fong2018net2vec,kim2018tcav,koh2020conceptbottleneck,zhou2018ibd} that provide an explanation for an output of a CNN model layer in terms of pre-defined semantic concepts. However, different from these works, UFO enables users to explicitly set the desired levels of faithfulness and understandability, and seamlessly obtain an explanation that is suitable for their needs. 

Our main contributions are as follows:
\begin{itemize}
    \item We operationalize the notions of ``faithfulness'' and ``understandability'' and propose a set of definitions for each of these terms. We then integrate them into a unified concept-based explanation method which is flexible enough to accommodate the different notions depending on the downstream application.
    \item We demonstrate how the method can be used to generate explanations of varying levels of faithfulness and understandability, and analyze where these explanations differ. This reveals a number of insights into how concept are selected by concept-based explanations, and which concepts are more likely to be used to explain a model. We demonstrate how frequency, size, and learnability of the selected concepts differ across the different explanation objectives.
    \item We discuss how our method generalizes and unifies most prior concept-based explanation approaches. We empirically validate this assertion by comparing our produced explanations to those from 3 prior works (NetDissect~\cite{bau2017netdissect}, IBD~\cite{zhou2018ibd} and ELUDE~\cite{ramaswamy2022elude}).  
\end{itemize}

UFO enables researchers and practitioners to reason concretely about choices for understandability and faithfulness, and compare to methods with similar incentives.

\section{Related work}
\label{sec:related}

Concept-based explanation methods are a popular class of model explanation methods that explain model behaviour and outputs with human-understandable semantic concepts. 
They can be categorized along several axes.
First, there are methods that explain some aspect of a model post-hoc~\cite{bau2017netdissect,fong2018net2vec,kim2018tcav,ramaswamy2022elude,zhou2018ibd}.
NetDissect~\cite{bau2017netdissect} and Net2Vec~\cite{fong2018net2vec} give insights about what the model has learned by identifying which neuron or which set of neurons encode a specific concept.
On the other hand, TCAV (Testing with Concept Activation Vectors)~\cite{kim2018tcav} learns vectors in the model activation space that corresponds to concepts and use them to quantify how sensitive the model's predictions are to a specific concept. Other methods focus on explaining how a model predicts a certain target class~\cite{ramaswamy2022elude,zhou2018ibd}.
IBD (Interpretable Basis Decomposition)~\cite{zhou2018ibd} first learns concept vectors in the model activation space, then decomposes the model's prediction in terms of these vectors. 
ELUDE (Explanation via Labelled and Unlabelled DEcomposition)~\cite{ramaswamy2022elude} also decomposes the model's prediction, but focuses on characterizing what portion of the prediction can and cannot be explained with available concepts.

More recently, concept-based ``interpretable-by-design'' models have been proposed~\cite{koh2020conceptbottleneck,marcos_accv_2020,radenovic2022neural,dubey2022scalable}.
They all take the concept bottleneck approach where the full model consists of a bottleneck that recognizes concepts in an input image and a discriminative model that classifies the image based on the bottleneck outputs (concept scores).
The discriminative model is parameterized as an interpretable model (e.g., linear model, generalized additive model), which makes the full model interpretable, i.e., an interpretable discriminative model reasoning with interpretable features (concepts recognized by the bottleneck).
The differences between each work lie in their choice of parameterization and training algorithm for the bottleneck and the discriminative model.

Although all these methods are called ``concept-based,''
the relationship between them are largely unclear.
This poses a huge challenge to researchers and practitioners who want to compare different methods' capabilities, constraints, and (implicit) assumptions.
In this work, we address this challenge by presenting a unified method that encapsulates and characterizes all aforementioned methods with respect to two axes: faithfulness and understandability.

Our work is similar in spirit to a growing body of works that introduce analyses and frameworks to better understand, evaluate, and compare model explanation methods~\cite{adebayo2018sanity,adebayo2022iclr,adebayo2020neurips,han2022LFA,kim2022hive,ramaswamy2023overlookedfactors,sokol2020explainability}.
In particular, Han et al.~\cite{han2022LFA} also argue the need for a unified framework and propose one for attribution heatmap explanations (specifically perturbation and gradient-based methods), which aim to explain which input features are relevant to a model's output decision.
Ramaswamy et al.~\cite{ramaswamy2023overlookedfactors} examine concept-based methods as we do, but they focus on analyzing commonly overlooked factors: probe dataset choice, concept learnability, and number of concepts used in an explanation.
On the other hand, we introduce a unified method (UFO) that deepens our understanding of the behavior of concept-based methods and their relation to one another.

UFO also enables researchers and practitioners to generate customized explanations by turning the knobs of faithfulness and understandability.

\section{UFO: A unified method for generating concept-based explanations for CNNs}
\label{sec:method}
We now propose our unified method (Understandability and Faithfulness Operationalization) for concept-based explanations, which explain some aspect of the model's output (either its final prediction or intermediate scores) in terms of a set of pre-defined semantic concepts. We develop this as a post-hoc explanation method: generating an explanation when given access to the trained model as well as a \emph{probe dataset}, a dataset similar to the training dataset that is labelled with a set of concepts. 

\smallsec{Given} Concretely, consider an image classification CNN model $F\colon\mathcal{X} \rightarrow \mathbb{R}^D$, which outputs a vector of dimension $D$ when given an image $x\in\mathcal{X}$ as an input, and let $y(x) = \argmax F(x)$, which outputs a model's prediction. 
Note that $F$ here may be the final or an intermediate layer output of the model, corresponding to the layer we aim to explain.\footnote{If $F$ is the final layer of the model, we will demonstrate in~\cref{ssec:operationalization} that our framework can be adapted to explain the model's predicted class $y(x) = \argmax F(x)$ instead of the model's full output.}

We want to explain this output in terms of $C$ semantic concepts. To do so, we use a probe dataset $X\subset \mathcal{X}$, where each $x \in X$ is annotated $A(x) \in \{0,1\}^C$ with the presence or absence of each of these concepts. By correlating the presence or absence of these concepts with the model's output, we can identify what human-understandable concepts are contributing to the model's predictions. 

\smallsec{Explanation framework} We consider $F=g \circ f$ as a combination of an intermediate function $f \colon \mathcal{X} \rightarrow \mathbb{R}^n$, which produces a set of $n$ image features (that we will attempt to explain using the $C$ concepts), and a function $g \colon \mathbb{R}^n \rightarrow \mathbb{R}^D$, which combines the features into the $D$-dimensional output.

We explain $F$ by learning two functions $h_{\text{conc}}\colon \mathbb{R}^n \rightarrow \mathbb{R}^C$, which maps the features $f$ to the concepts, and a function $h_\text{pred}\colon\mathbb{R}^C \rightarrow \mathbb{R}^D$, which maps the concepts to the model output $F$. Then, we have two different objectives: 
\begin{enumerate}
    \item To maximize the \textbf{faithfulness} of the explanation. Simply put, this objective requires the explanation to mimic the model's output as far as possible, i.e., $F(x) \approx h_\text{pred} \circ h_{\text{conc}} \circ f(x)$. 
    \item To maximize the \textbf{understandability} of the explanation. An explanation that perfectly mimics the model but is not human-understandable doesn't make the model more interpretable: humans need to be able to parse the explanation resulting from $h_\text{pred}$ and $h_{\text{conc}}$. 
\end{enumerate}

Currently, there is no clear consensus on what these two different terms could mean. In the following subsections, we explore the definitions that each of these objectives could take and highlight how our proposed definitions can describe existing concept-based explanations. 

\subsection{Faithfulness} 
We can vary the faithfulness of an explanation by changing the definition for how an explanation should mimic a model's output (i.e., how an explanation of the form $h_{\text{pred}} \circ h_{\text{conc}}$ approximates $g$, the latter part of a model). We describe three definitions below:
\begin{enumerate}
\itemsep0em
\item \textbf{Most faithful (\MF).} 
First, we can match the full $D$-dimensional output of our explanation with that of the model
for all possible images, not just those in the probe dataset (i.e., $\mathcal{X}$ instead of $X$).
One way to achieve this would be to learn $h_\text{conc}$ and $h_\text{pred}$ such that $h_\text{pred} \circ h_\text{conc} = g$. 

\item \textbf{\textcolor{black}{Somewhat faithful} (\SF).}
Next, we can relax our first definition and require that the full outputs of the explanation and model match only for images in the probe dataset.
Then, rather than requiring $h_\text{pred} \circ h_\text{conc} = g$, we would minimize the following mean-squared error (MSE) loss for all $x \in X$:
\begin{equation}
    \| g \circ f(x) - h_\text{pred} \circ h_\text{conc} \circ f(x)\|.
\end{equation}
This would be potentially more useful to developers who wish to debug or improve a model: explanations that model the full score distribution allow for better diagnoses of the model. 

\item \textbf{\textcolor{black}{Least faithful} (\LF).} 
Finally, instead of mimicking the model's full output, our explanation can mimic just the model's prediction.
Then, we would minimize the following cross-entropy (CE) loss for all $x \in X$: 
\begin{equation}
    CE(y(x),  h_\text{pred} \circ h_\text{conc}).
\end{equation}
This would be more useful for end-users of the model, who just want to understand how a specific prediction is being made. 
\end{enumerate} 

\subsection{Understandability}
\label{subsec:understandability}
The understandability of an explanation can vary along the following three axes:

\smallsec{Complexity of functions} The choice of the two learned explanation functions $h_\text{conc}$ and $h_\text{pred}$ can vary the understandability of the explanation.
These can be general functions; however, they are typically chosen as linear functions in most concept-based explanation methods.
A linear $h_\text{conc}$ makes it easy to learn $h_\text{conc}$, while a linear $h_\text{pred}$ makes it easy for a human to understand a model's prediction as a linear combination of concepts.

\smallsec{Number of concepts} 
    Prior works have shown that humans can realistically only reason with a small number of $K$ concepts (typically with $K=16$ or $32$)~\cite{ramaswamy2023overlookedfactors}. 
    Thus, we allow the user to select $K$ concepts using a selection matrix $S$ that picks $K$ out of $C$ concepts and let $\mathbbm{1}_S$ be the indicator vector of these $K$ chosen concepts. Then, we can describe our explanation as $h_\text{pred}  \circ \mathbbm{1}_S \circ h_\text{conc}\circ f(x)$.

\smallsec{Concept encoding} 
    Humans can more easily reason with binary values for concepts (e.g., is a bed present or absent?) than they can with continuous values for concepts (e.g., the probability of bed being present is 0.74). 
    Thus, we can describe how concepts are encoded as binary values (\MU), probabilities (\SU), or continuous values (\LU).
    When given a concept encoding $u \in \mathcal{C}$, we define $\mathcal{C}$ as follows:
    \begin{enumerate}[noitemsep,topsep=0pt]
    \item \textbf{\textcolor{black}{Most understandable} (\MU).} $\mathcal{C} = \{0,1\}^C$.
    \item \textbf{\textcolor{black}{Somewhat understandable} (\SU).} $\mathcal{C} = [0,1]^C$.
    \item \textbf{\textcolor{black}{Least understandable} (\LU).} $\mathcal{C} = \mathbb{R}^C$.
    \end{enumerate}
    
In our main experiments, we use linear functions for $h_\text{conc}$ and $h_\text{pred}$, set $K = 16$ concepts, and vary understandability based on how concepts are encoded.

\subsection{Operationalization of explanations}
\label{ssec:operationalization}
Now, we instantiate specific objective functions using our different definitions of faithfulness and understandability.
Our objective functions are all of the following form, where we find the optimal selection matrix $S$ that chooses $K$ semantic concepts, and the optimal functions $h_\text{conc}$ mapping the features $f$ to the $C$ semantic concepts and $h_\text{pred}$ mapping these concepts to the model predictions:
\begin{align}
    \argmin_{h_\text{conc}, h_\text{pred}, S} \lambda_1 L_{\text{mimic}} + \lambda_2 L_{\text{align}}
\end{align}
Here, $L_\text{mimic}$ varies based on the specific definitions used and describes how an explanation should mimic a model, while $L_\text{align}$ is fixed as follows and describes how $h_\text{conc}$ aligns features to concepts for images in the probe dataset:
\begin{align}
    L_\text{align} = \sum_{x \in X} \sum_i CE(\mathbbm{1}_S\circ h_\text{conc}\circ f(x)_i, \mathbbm{1}_S \circ A_i(x))
\end{align}
Then, hyperparameters $\lambda_1$ and $\lambda_2$ allow us to prioritize the mapping from the features to the concepts over the mapping from the concepts to the final output, and vice-versa.

We start with the simplest definition of $L_\text{mimic}$.

\noindent\textbf{Least understandable}, \textbf{\textcolor{black}{most faithful} (\LU, \MF)}:
\begin{equation}
\label{eq:mf}
L_\text{mimic}^\text{\LU, \MF} = 
   \sum_{x \in \mathcal{X}}\|g \circ f(x) - h_\text{pred} \circ \mathbbm{1}_S \circ h_\text{conc}\circ f(x)\|
\end{equation}  
However, decomposing a general $g$ into $h_\text{conc}$ and $h_\text{pred}$ is not tractable, as this would require us to minimize the above for all images $x \in \mathcal{X}$ (not just those in the probe dataset $X$).

Instead, we consider only losses that are tractable by using less strict definitions of faithfulness below.\footnote{Changes to the previous equation are denoted in \textbf{\textcolor{red}{bolded red}}.}  

\noindent {\bf \textcolor{black}{Least understandable}, \textcolor{black}{somewhat faithful}  (\LU, \SF):} 
\begin{equation}
\label{eq:sf_lu}
   L_\text{mimic}^\text{\LU, \SF} = \sum_{\textcolor{red}{\boldsymbol{x \in X}}} \|g\circ f(x) - h_\text{pred}  \circ \mathbbm{1}_S \circ h_\text{conc}\circ f(x)\| 
\end{equation}

\noindent Here, we are only concerned mimicking full outputs on the probe dataset and thus have a tractable objective.

\cref{eq:sf_lu} can also be modified to be more understandable.
Rather than using the continuous output of $h_\text{conc} \circ f(x)$, we could also use a probabilistic version of it.

\noindent {\bf \textcolor{black}{Somewhat understandable}, \textcolor{black}{somewhat faithful} \ (\SU, \SF):} 
\begin{equation}
\label{eq:sf_su}
   L_\text{mimic}^\text{\SU, \SF} = 
   \sum_{x \in X} \|g \circ f(x) - h_\text{pred}  \circ \mathbbm{1}_S \circ 
   \textcolor{red}{\boldsymbol{p}} \circ h_\text{conc}\circ f(x)
   \|
\end{equation}
where $p \colon \mathbb{R}^n \to [0, 1]^C$ is a function that maps features to probabilities (e.g. the sigmoid function).

We can make the explanation even more understandable by replacing $p \circ h_\text{conc}\circ f(x)$ entirely with the binary, ground-truth attributes encoded by $A(\cdot)$, i.e., explaining the model’s output with perfect knowledge of the concepts.

\noindent {\bf \textcolor{black}{Most understandable}, \textcolor{black}{somewhat faithful}, (\MU, \SF):} 
\begin{equation}
\label{eq:sf_mu}
   L_\text{mimic}^\text{\MU, \SF} = \sum_{x \in X} \|g \circ f(x) - h_\text{pred}  \circ \mathbbm{1}_S \circ \textcolor{red}{\boldsymbol{A(x)}}\|
\end{equation}

Finally, we could use the least strict definition of faithfulness, where we only care about mimicking the single model prediction $y(x)$ rather than its full output $g \circ f(x)$. This can be paired with all three understandability definitions.

\noindent {\bf \textcolor{black}{Least understandable}, \textcolor{black}{least faithful}  (\LU, \LF):}\footnote{Changes to $L_\text{mimic}^\text{\LU, \SF}$ given by \cref{eq:sf_lu} are denoted in \textcolor{NavyBlue}{\textbf{bolded blue}}.}
\begin{equation}
\label{eq:lf_lu}
   L_\text{mimic}^\text{\LU, \LF} = \sum_{x \in X} \textcolor{NavyBlue}{\boldsymbol{CE}}(y(x),h_\text{pred}  \circ \mathbbm{1}_S \circ h_\text{conc}\circ f(x))
\end{equation}

\noindent {\bf Somewhat understandable, \textcolor{black}{Least faithful} (\SU, \LF):} 
\begin{equation}
\label{eq:lf_su}
   L_\text{mimic}^\text{\SU, \LF} = \sum_{x \in X} CE(y(x) , h_\text{pred}  \circ \mathbbm{1}_S \circ 
   \textcolor{red}{\boldsymbol{p}} \circ h_\text{conc}\circ f(x) )
\end{equation}

\noindent {\bf Most understandable, \textcolor{black}{least faithful}(\MU, \LF):} 
\begin{equation}
\label{eq:lf_mu}
   L_\text{mimic}^\text{\MU, \LF} = \sum_{x \in X} CE (y(x) , h_\text{pred} \circ \mathbbm{1}_S \circ \textcolor{red}{\boldsymbol{A(x)}})
\end{equation}

\subsection{Optimization}
The trickiest part of the optimization is the selection of $K$ concepts out of $C$ concepts in total, since this is non-differentiable. Based on prior work, we assume $h_\text{pred}$ and $h_\text{conc}$ to be linear functions, and use a group Lasso regularization~\cite{yuan2006model} that forces the squared sum of the coefficients of a concept to 0. That is, $h_\text{pred}$ is learned as a coefficient matrix $W_\text{pred} \in \mathbb{R}^{D \times C}$. Assuming that the columns of $W_\text{pred}$ are sorted in increasing order of their squared $\ell_2$ norm ($\sum_{j =1}^D (W_\text{pred})_{j, i}^2$) (i.e. 1st column has smallest $\ell_2$ norm, followed by 2nd column, etc.), during training, we add the following regularization loss: 
\begin{equation}
\label{eq:regularization}
L_\text{reg} = \sum_{i=1}^{C-K} \Big|\sum_{j =1}^D (W_\text{pred})_{j, i}^2 \Big|
\end{equation}

We gradually increase the weight of this loss during training, initially allowing the $W_\text{pred}$ to use all concepts to identify the most relevant ones, and then forcing the $C - K$ smallest columns of $W_\text{pred}$ to 0.

\section{Experiments}
\label{sec:experiments}

\begin{table}[t]
\centering
\begin{tabular}{c|cc|cc|cc}
\toprule
\multicolumn{1}{l|}{} & \multicolumn{2}{c|}{Binary} & \multicolumn{2}{c|}{Grouped} & \multicolumn{2}{c}{Fine-grained} \\
\midrule
\multicolumn{1}{l|}{} & \SF & \LF & \SF & \LF & \SF & \LF \\
\midrule
\MU & 4.20 & 7.13 & 2.95 & 12.59 & 4.12 & 60.43 \\

\SU & 6.79 & 5.00 & 2.54 & 7.81 & 3.39 & 41.26 \\
\LU & \textbf{0.93} & 15.08 & \textbf{1.26} & 9.81 & \textbf{2.20} & 23.80 \\\bottomrule
\end{tabular}
\caption{\textbf{Gap between explanation and model outputs (\cref{sssec:faithfulness}).}
For three Places365 models (binary, grouped, and fine-grained), we report the mean L2 distance between the distributions output by the explanation and the model ($\downarrow$ is better) when varying levels of faithfulness and understandability.
For each model, we \textbf{bold} the most faithful explanation, i.e., explanation with the lowest mean L2 distance. 
As expected, somewhat faithful (\SF) explanations have lower mean L2 distance than least faithful (\LF) explanations.
Also as expected, least understandable (\LU) explanations have lower mean L2 distance than somewhat understandable (\SU) and most understandable (\MU) explanations, demonstrating a faithfulness-vs-understandability tradeoff.
}
\label{tab:faithful}
\end{table}

In~\cref{ssec:experiments_tradeoff}, we analyze the faithfulness-vs-understandability tradeoff by optimizing \cref{eq:sf_lu,eq:sf_su,eq:sf_mu,eq:lf_lu,eq:lf_su,eq:lf_mu} 
for the same model and examine which concepts are highlighted in each case. We find that an explanation's accuracy decreases as the explanation is made more understandable and that the concepts highlighted are highly dependent on the equation optimized. 
In~\cref{ssec:experiment_prior_work}, we compare our results to that of prior works~\cite{bau2017netdissect,zhou2018ibd,ramaswamy2022elude}. 

\subsection{Faithfulness-vs-understandability tradeoff}
\label{ssec:experiments_tradeoff}

\smallsec{Experimental setup}
\label{ssec:implementation_details}
We use a ResNet18~\cite{he2016resnet} model trained on Places365~\cite{zhou2017places} as our blackbox model to explain. This model takes as input an image and outputs a vector with the predicted probabilities of the image belonging to each of the 365 classes. Similar to ELUDE~\cite{ramaswamy2022elude}, we consider these predictions at 3 different granularities (using annotations provided in the dataset):

\begin{enumerate}[noitemsep,topsep=0pt]
    \item \textbf{Binary} (2 classes): ``indoor'' vs. ``outdoor'' scenes
    \item \textbf{Grouped} (16 classes): coarse-grain scene categories (e.g., ``home/hotel'' or ``forest/field/jungle'')
    \item \textbf{Fine-grained} (365 classes): scene labels (e.g., ``bedroom'' or ``bamboo forest'')
\end{enumerate}

For binary and grouped scene predictions, we replace and retrain the final layer of the model. For fine-grained predictions, we analyze the 365-class model but focus only on explaining the top 20 classes that are most represented within the probe dataset, to simplify computation.

We use the ADE20k~\cite{zhou2017ade20k,zhou2019ade20k_ijcv} dataset (license: BSD 3-Clause) as the probe dataset with which to generate explanations, splitting its images randomly into train (60\%, 11839 images), val (20\%, 3947 images), and test (20\%, 3947 images).
We use train to optimize~\cref{eq:sf_lu,eq:sf_su,eq:sf_mu,eq:lf_lu,eq:lf_su,eq:lf_mu},
val to pick hyperparameters (e.g., $\lambda_1, \lambda_2$), and test to measure the accuracy of the explanation (i.e., the number of images for which the discrete explanation output matches that of the model).
In~\cite{bau2017netdissect}, this dataset was further densely labelled with 1197 concepts comprising of objects, object parts, scenes, colors and textures; we use a subset of these concepts. First, we remove concepts that occur in fewer than 20 images within the training dataset. This gives us a set of 309 concepts. We further prune these concepts to remove concepts that are correlated with each other, following the findings of~\cite{ramaswamy2023overlookedfactors}. 
Correlations between concepts can be computed using the ground truth scores or the learned score of the concept. Surprisingly, we find that these result in very different correlations. Thus, we select the set of concepts separately for each setting of understandability (\MU, \SU, \LU). More details are given in the appendix.

\begin{table*}[t]
    \centering
   
    \begin{tabular}{p{2cm} l  L{4cm}L{4cm}  L{4cm}}
 \toprule
& & \multicolumn{1}{c}{most understandable} & \multicolumn{1}{c}{somewhat understandable} & \multicolumn{1}{c}{least understandable}  \\ 
& & \multicolumn{1}{c}{\MU} & \multicolumn{1}{c}{\SU} & \multicolumn{1}{c}{\LU} \\
\midrule 
bedroom & \SF & \textcolor{NavyBlue}{bed}, \textcolor{red}{\textit{building}}, \textcolor{NavyBlue}{pillow} & \textcolor{red}{\textit{bulletin board}}, \textcolor{red}{\textit{\textbf{counter}}}, \textcolor{red}{\textit{footbridge}} & \textcolor{NavyBlue}{\textbf{wardrobe}}, \textcolor{red}{\textit{pitcher}}, \textcolor{red}{\textit{flag}}\\
 & \LF & \textcolor{red}{\textit{cap}}, \textcolor{red}{\textit{buffet}}, \textcolor{red}{\textit{saucepan}} & \textcolor{red}{\textit{\textbf{counter}}}, \textcolor{red}{\textit{food}}, \textcolor{red}{\textit{refrigerator}} & \textcolor{NavyBlue}{clock}, \textcolor{NavyBlue}{lamp}, \textcolor{NavyBlue}{\textbf{wardrobe}}\\\midrule
conference-room & \SF & \textcolor{red}{\textit{\textbf{bed}}}, \textcolor{NavyBlue}{\textbf{table}}, \textcolor{red}{\textit{building}} & \textcolor{NavyBlue}{loudspeaker}, \textcolor{red}{\textit{faucet}}, \textcolor{NavyBlue}{table tennis} & \textcolor{NavyBlue}{double door}, \textcolor{NavyBlue}{\textbf{table}}, \textcolor{NavyBlue}{desk}\\
 & \LF & \textcolor{red}{\textit{\textbf{bed}}}, \textcolor{red}{\textit{rack}}, \textcolor{red}{\textit{cup}} & \textcolor{red}{\textit{tree}}, \textcolor{red}{\textit{earth}}, \textcolor{red}{\textit{sink}} & \textcolor{NavyBlue}{\textbf{table}}, \textcolor{NavyBlue}{crt screen}, \textcolor{NavyBlue}{plant}\\\midrule
crosswalk & \SF & \textcolor{NavyBlue}{road}, \textcolor{NavyBlue}{building}, \textcolor{NavyBlue}{person} & \textcolor{NavyBlue}{\textbf{backpack}}, \textcolor{red}{\textit{\textbf{land}}}, \textcolor{red}{\textit{cap}} & \textcolor{NavyBlue}{\textbf{backpack}}, \textcolor{red}{\textit{paper}}, \textcolor{NavyBlue}{flag}\\
 & \LF & \textcolor{red}{\textit{step}}, \textcolor{red}{\textit{floor}}, \textcolor{red}{\textit{platform}} & \textcolor{red}{\textit{bush}}, \textcolor{red}{\textit{\textbf{land}}}, \textcolor{red}{\textit{loudspeaker}} & \textcolor{NavyBlue}{\textbf{backpack}}, \textcolor{NavyBlue}{poster}, \textcolor{red}{\textit{mountain}}\\\midrule
downtown & \SF & \textcolor{NavyBlue}{building}, \textcolor{red}{\textit{\textbf{wall}}}, \textcolor{NavyBlue}{sky} & \textcolor{NavyBlue}{telephone booth}, \textcolor{NavyBlue}{footbridge}, \textcolor{NavyBlue}{backpack} & \textcolor{NavyBlue}{baby buggy}, \textcolor{NavyBlue}{\textbf{flag}}, \textcolor{red}{\textit{ladder}}\\
 & \LF & \textcolor{red}{\textit{rack}}, \textcolor{red}{\textit{step}}, \textcolor{NavyBlue}{platform} & \textcolor{red}{\textit{bush}}, \textcolor{red}{\textit{seat}}, \textcolor{red}{\textit{loudspeaker}} & \textcolor{NavyBlue}{\textbf{flag}}, \textcolor{red}{\textit{door}}, \textcolor{red}{\textit{\textbf{wall}}}\\\midrule
kitchen & \SF & \textcolor{NavyBlue}{work surface}, \textcolor{red}{\textit{\textbf{bed}}}, \textcolor{red}{\textit{building}} & \textcolor{red}{\textit{grandstand}}, \textcolor{red}{\textit{backpack}}, \textcolor{NavyBlue}{\textbf{faucet}} & \textcolor{NavyBlue}{pitcher}, \textcolor{NavyBlue}{crt screen}, \textcolor{NavyBlue}{\textbf{faucet}}\\
 & \LF & \textcolor{red}{\textit{\textbf{bed}}}, \textcolor{red}{\textit{\textbf{loudspeaker}}}, \textcolor{red}{\textit{post}} & \textcolor{red}{\textit{\textbf{loudspeaker}}}, \textcolor{red}{\textit{sand}}, \textcolor{NavyBlue}{\textbf{refrigerator}} & \textcolor{NavyBlue}{cabinet}, \textcolor{NavyBlue}{\textbf{refrigerator}}, \textcolor{NavyBlue}{windowpane}\\\midrule
\end{tabular}
\caption{\textbf{Selected concepts vary based on faithfulness-understandability setting.} 
We compare the concepts selected to explain the fine-grained scene model under the six faithfulness-understandability settings (\{\MU, \SU, \LU\} $\times$ \{\SF, \LF\}). We focus on a subset of randomly-selected scenes. For each, we highlight the top 3 concepts (i.e., highest absolute coefficients of $h_{pred}$). \textcolor{red}{\textit{Red}} denotes a concept with a negative coefficient, \textcolor{NavyBlue}{blue} denotes one with a positive coefficient, and \textbf{bold} emphasize a concept that is a top-3 concept for at least two of the six explanations for a given scene. We find that the concepts highlighted differ wildly across different definitions of faithfulness and understandability.
}
    \label{tab:finegrained_attr}
\end{table*}

We use linear models for $h_\text{conc}$ and $h_\text{pred}$, with $f(x)$ as the output of the penultimate layer of the model. These are trained for 5000 epochs with batch size = 1024 using either cross entropy or MSE loss (F vs. FF) and the Adam~\cite{KB14Adam} optimizer with a learning rate = 1e-3. We set $\lambda_1$ and $\lambda_2$ as hyperparameters by picking the ratio $\lambda_1 \colon \lambda_2$ that has the highest validation accuracy.
Finally, we set the number of concepts used to $K = 16$ (for the binary classifier) or $K=32$ (for the grouped and fine-grained classifiers) concepts, based on prior work~\cite{ramaswamy2023overlookedfactors}. See the supp. mat. for details.

\subsubsection{How well do explanations emulate the model?}
\label{sssec:faithfulness}
We first analyze how well different explanations emulate the model as we vary the levels of faithfulness and understandability.
We do so by measuring and comparing the average L2 distance between the distributions output by the explanation and the model, which capture how well the explanation emulate the model. See Tab.~\ref{tab:faithful} for the full results.
We find that, as expected, the L2 distance is lower for the somewhat faithful (\SF) explanation, as compared to the least faithful (\LF) explanation. Moreover, we find that as the understandability increases, the faithfulness decreases:
the least understandable (\LU) explanation emulates the model better than the somewhat understandable (\SU) and the most understandable (\MU) explanations. This suggests a trade-off between faithfulness and interpretability: explanations that emulate the model well can be less understandable. Applications that require precise mimicking of the model's predictions (for example, if being used to debug a model) would be better explained using the less understandable, more faithful (\LU, \SF) version of UFO, whereas explanations being given to a lay-person might require higher understandability and thus, be better explained using a more understandable, less faithful (\MU, \LF) version of UFO. 

\subsubsection{How do ``important'' concepts change?}
\label{sssec:understandability}
Next we analyze how ``important'' concepts, i.e., concepts selected by the explanation, change as we vary the faithfulness and understandability objectives.
We find that the important concepts are very different under different objectives: 
~\Cref{tab:finegrained_attr} shows the three most important concepts (i.e., the three concepts with the highest absolute coefficients in $h_{pred}$) for a subset of coarse-grained scene groups.

\begin{figure}[t]
    \centering
   \includegraphics[width=\linewidth]{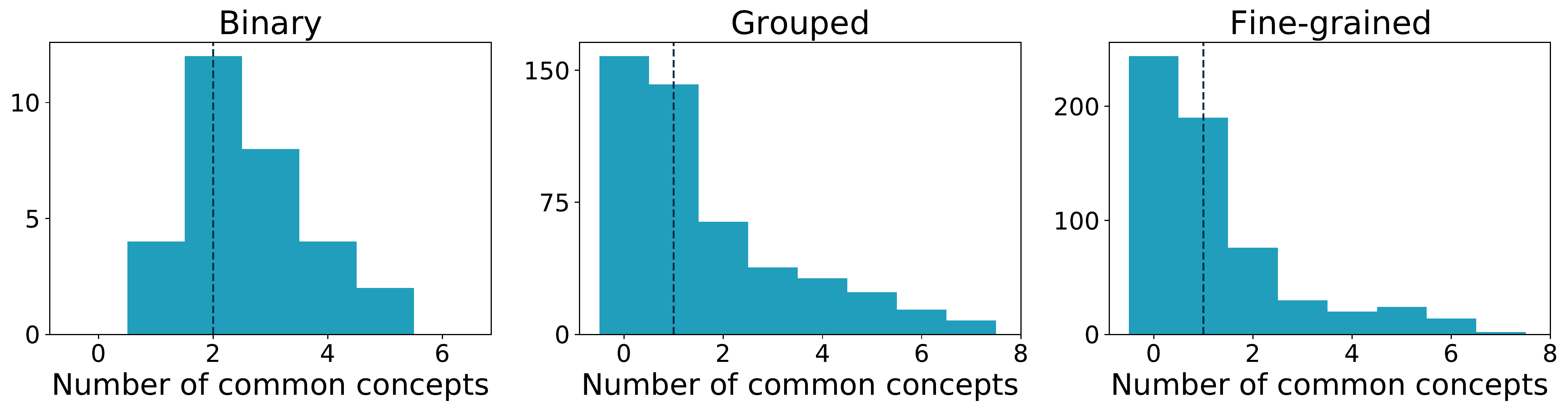}
    \caption{
    \textbf{Concept overlap (\cref{sssec:understandability})}.
    For three Places365 models (binary, grouped, and fine-grained), we report the number of shared concepts selected by the six different types of explanations (\{\MU, \SU, \LU\} $\times$ \{\SF, \LF\}). Concretely, for each scene class, we identify the 10 most important concepts (based on the absolute coefficient value) for each type of explanation, and measure the overlap in concepts between each pair of explanations.
    We see that the average overlap is extremely low, between 1 and 2 concepts for most classes, indicating that the selection of the right understandability and faithfulness objectives is critical.}
    \label{fig:conc_hist}
\end{figure}

We notice there are significant differences between the selected concepts under different settings of faithfulness and understandability. For example, consider the most understandable explanation setting. When using the somewhat faithful (\SF) setting, the top 3 concepts for the scene ``bedroom'' include ``bed,'' ``building'' and ``pillow'' (with a negative coefficient for ``building''). For the least faithful (\LF) setting, the top 3 concepts are instead ``cap'', ``buffet'', and ``saucepan.'' 
In order to quantify this better, we consider the overlap among the top 10 concepts for each class. We see that the median overlap is just 1 for fine-grained scenes and grouped scenes, and 2 for binary (Fig.~\ref{fig:conc_hist}).

\begin{figure}[t]
    \centering
    \includegraphics[width=\linewidth]{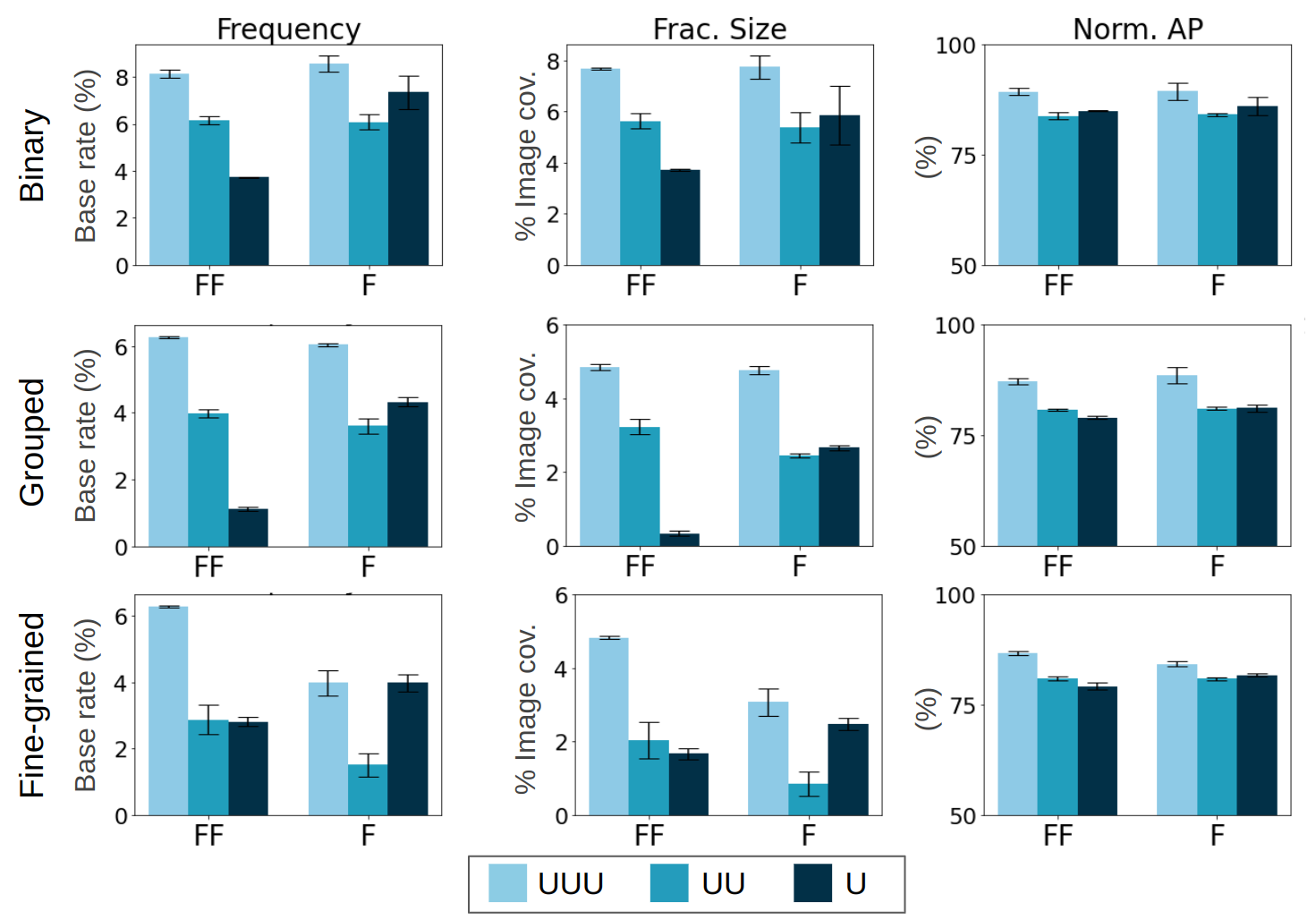}    
    \caption{\textbf{Comparison of concepts chosen}. We compare the concepts based on their \textit{frequency}, the average \textit{size} of the concept within an image (fraction size) and the \textit{learnability} of the concept from the feature space (normalized AP), across the 6 ($\{$\MU, \SU, \LU $\} \times \{$\SF, \LF$\}$) settings. In general, we see that concepts chosen by the \MU{} setting are typically larger, occur more often and are easier to learn, compared to the \SU{} and \LU{} settings. }
    \label{fig:understand}
\end{figure}

Given that the concept overlap is so low, we analyze what concepts are chosen by different types of explanations. We consider different aspects of the concepts chosen:  the frequency (fraction of images that contain the concept), the average size (fraction of the image occupied by the segmentation mask of the concept), and the learnability (measured using the normalized average precision and the ROC AUC). In general, we find that concepts highlighted by the most understandable (\MU) explanation tend to occur more often, be larger in size,  and be more learnable, as shown in Fig.~\ref{fig:understand}. For somewhat faithful explanations (\SF), we see that for all metrics, somewhat understandable (\SU) explanations contain more frequently occurring, larger and easier to learn concepts than less understandable (\LU) explanations. However, for less faithful (\LF) explanations, 
this trend does not always hold. We interpret this as follows.

For the \MU{} formulation, the \textit{frequency} of the concept occurrence is directly related to the amount of information encoded within the ground-truth concept annotations. Thus, concepts with higher base rates are more likely to be used within the explanation.
For \SU{} and \LU{} formulations, this is not the case, since we use either the learned scores for the concepts (either $h_\text{conc} \circ f(x)$ or a probabilistic version of it), and these continuous vectors contain information regardless of the base rates of the concepts. 
Next, the \textit{size} of the concept can influence its \textit{learnability}: larger concepts can be easier for a model to learn, thus, these concepts are more likely to be used to learn a scene class.
However, for the \SU{} and \LU{} formulations, we note that the concept encodings could include other information within it (potentially even additional unlabelled concepts). Thus, the selected concepts might be the ones that are themselves less learnable.

\subsection{Comparison to prior work}
\label{ssec:experiment_prior_work}
Finally, we consider prior works that generate concept-based explanations and analyze them within our framework. We find that we are able to express all concept-based explanations~\cite{bau2017netdissect,fong2018net2vec,kim2018tcav,zhou2018ibd,koh2020conceptbottleneck,ramaswamy2022elude} in our framework (see supp. mat. for more details). For methods whose optimizations are slightly from ours~\cite{bau2017netdissect,zhou2018ibd,ramaswamy2022elude}, we also run the closest form of our optimization along with their method, and compare the results produced. 

\smallsec{NetDissect~\cite{bau2017netdissect}} 
Rather than explaining the full output of the network $F(x) \in \mathbb{R}^D$, NetDissect explains $F(x) \in \mathbb{R}$ for each neuron. Furthermore, NetDissect correlates the neuron's output with the pixel-level \emph{segmentation} of each concept so the formulation changes slightly to accommodate predicting concepts at an image-level to a pixel-level. Generally, this can be achieved with $\lambda_1 = 0$ and changing $L_{\text{align}}$ to denote how well aligned an individual neuron's output is to concept segmentations.
Within our optimization, we set $\lambda_1 = 0$ and optimize $h_\text{conc}$ to find the best semantic concept for each neuron. 
While this is different to the segmentation intersection over union (IOU) score that NetDissect uses,  we still identify similar explanations (Tab.~\ref{tab:netdissect_comp}).

\begin{table*}[t]
\centering
\begin{tabular}{c|lc|lclclc}
\toprule
\multirow{2}{*}{neuron} & \multicolumn{2}{c|}{NetDissect~\cite{bau2017netdissect}} & \multicolumn{6}{c}{UFO (top 3 concepts)} \\
\cmidrule(l{3pt}r{3pt}){2-3} \cmidrule(l{3pt}r{3pt}){4-9}
& top concept & \multicolumn{1}{c|}{IOU} & concept-1 & \multicolumn{1}{c}{nAP-1} & concept-2 & \multicolumn{1}{c}{nAP-2} & concept-3 & \multicolumn{1}{c}{nAP-3} \\ \midrule
454 & car & 0.2184 & \textbf{car} & 94.4 & saddle & 94.0 & van & 92.2 \\
193 & skyscraper & 0.2055 & water tower & 99.2 & \textbf{skyscraper} & 96.8 & control tower & 95.3 \\
445 & car & 0.2014 & saddle & 99.0 & \textbf{car} & 96.1 & teapot & 95.8 \\
446 & pool table & 0.1928 & \textbf{pool table} & 97.8 & table tennis & 96.6 & arcade machine & 96.2 \\
500 & sofa & 0.1558 & \textbf{sofa} & 94.2 & bed & 93.0 & armchair & 91.4 \\
46 & house & 0.1549 & \textbf{house} & 96.1 & pavilion & 94.8 & windmill & 92.0 \\
341 & sea & 0.1531 & \textbf{sea} & 98.0 & lighthouse & 93.0 & ship & 91.1 \\
43 & bed & 0.1509 & \textbf{bed} & 96.9 & pillow & 96.8 & eiderdown & 96.0 \\
484 & water & 0.1496 & \textbf{water} & 92.0 & sea & 91.3 & river & 90.3 \\
329 & pool table & 0.1474 & \textbf{pool table} & 97.6 & table tennis & 97.0 & court & 94.8 \\
\bottomrule
\end{tabular}
\caption{\textbf{Comparison with NetDissect.} 
We show results for the 10 neurons that have the highest IOU scores with a concept in NetDissect~\cite{bau2017netdissect} alongside the top 3 concepts that best explain those units (i.e., highest normalized AP [nAP]) using our formulation.
Despite slight differences in the optimizations, we see that the concepts correspond well: the top concept from our explanation matches that of NetDissect for most units.
Concepts from our optimization that match the top concept from NetDissect are \textbf{bolded}.
}
\label{tab:netdissect_comp}
\end{table*}

Concretely, as in NetDissect, we set $f$ to the output of the final convolutional layer in the ResNet18~\cite{he2016resnet}, which outputs a $7 \times 7$ feature map for each neuron. ADE20k~\cite{zhou2017ade20k,zhou2019ade20k_ijcv} is labelled with segmentation masks. Thus, for each of the 49 regions, we are able to identify the most common object (or object part) and use this as a coarse segmentation map. We compute 
alignment using normalized AP~\cite{hoiem2012error} between the coarse segmentation map and the output of the neuron.
Other implementation details remain the same as before.

\smallsec{Net2Vec~\cite{fong2018net2vec} and TCAV~\cite{kim2018tcav}} Similar to NetDissect~\cite{bau2017netdissect}, these methods aim to explain the output of an intermediate layer. 
Here, $h_\text{conc}$ is a linear function that finds the best mapping from the feature space $f$ for each concept, and our optimization is exactly the same as that of these methods. 

\smallsec{IBD~\cite{zhou2018ibd}}
Here, the method attempts to explain the logits of the model. It assumes that both $h_\text{conc}$ and $h_\text{pred}$ are linear functions, and that the feature space $f$ is the output of the final convolutional layer. IBD optimizes \cref{eq:sf_su}, and thus is a (\SU, \SF) method. They optimize this equation by first computing $h_\text{conc}$ as a sequence of orthogonal vectors (called an ``interpretable basis''). They then express the model output as a linear combination of these basis vectors, with each weight being positive. They also limit the number of concepts per target class. This can be modelled by optimizing $S$ and $h_\text{pred}$ per target class.
Here, the main difference is that IBD adds a non-negative constraint to the coefficients in $h_\text{pred}$. This constraint appears to significantly affect concept selection, such that the set of concepts selected using the constraint vs. without it do not overlap in general. (results in the supp. mat.) 

\smallsec{Concept Bottleneck models~\cite{koh2020conceptbottleneck,marcos_accv_2020,radenovic2022neural,dubey2022scalable}} 
Unlike the other methods, Concept Bottleneck (CB) models are ``interpretable-by-design''. These models are learned as a composition of two functions, one that maps images to concepts and another that maps concepts to the final output. We focus on one, the concept bottleneck model~\cite{koh2020conceptbottleneck}, but others are similar. In our framework, this is an (\LU, \SF) method, i.e, it optimizes~\cref{eq:sf_lu}\footnote{Koh et al.~\cite{koh2020conceptbottleneck} also try adding a sigmoid layer after predicting concepts (i.e, optimizing Eq.~\ref{eq:sf_su}), but found a significant drop in model accuracy.}, where $h_\text{conc}$ is the identity function and $h_\text{pred}$ is $g$ itself. All concepts are allowed to be used within the explanation. However, the explanations generated can be hard to understand given the large number of concepts used, and the continuous encoding of the concepts. Removing the constraint of the number of concepts selected make the optimizations between Concept Bottleneck and the (\LU, \SF) identical, and hence, we do not run experiments. 

\smallsec{ELUDE~\cite{ramaswamy2022elude}}
In this work, the authors ignore the mapping from the features to the concepts entirely, with $\lambda_2 = 0$. They use the least faithful (\LF) and most understandable (\MU) definition and optimize \cref{eq:lf_mu}. Rather than pre-deciding $K$, they add a regularization constraint when optimizing to use fewer concepts. 
The main difference between the optimization (\MU, \LF) and ELUDE is in its L1 penalty. 
ELUDE enforces that the number of concepts used per target class is minimized using an L1 penalty (the total number of concepts used might be large), whereas we use a grouped L1 penalty to minimize the total number of concepts used.

\begin{figure}[t]
    \centering
    \vspace{-0.2in}
    \includegraphics[width=0.8\linewidth]{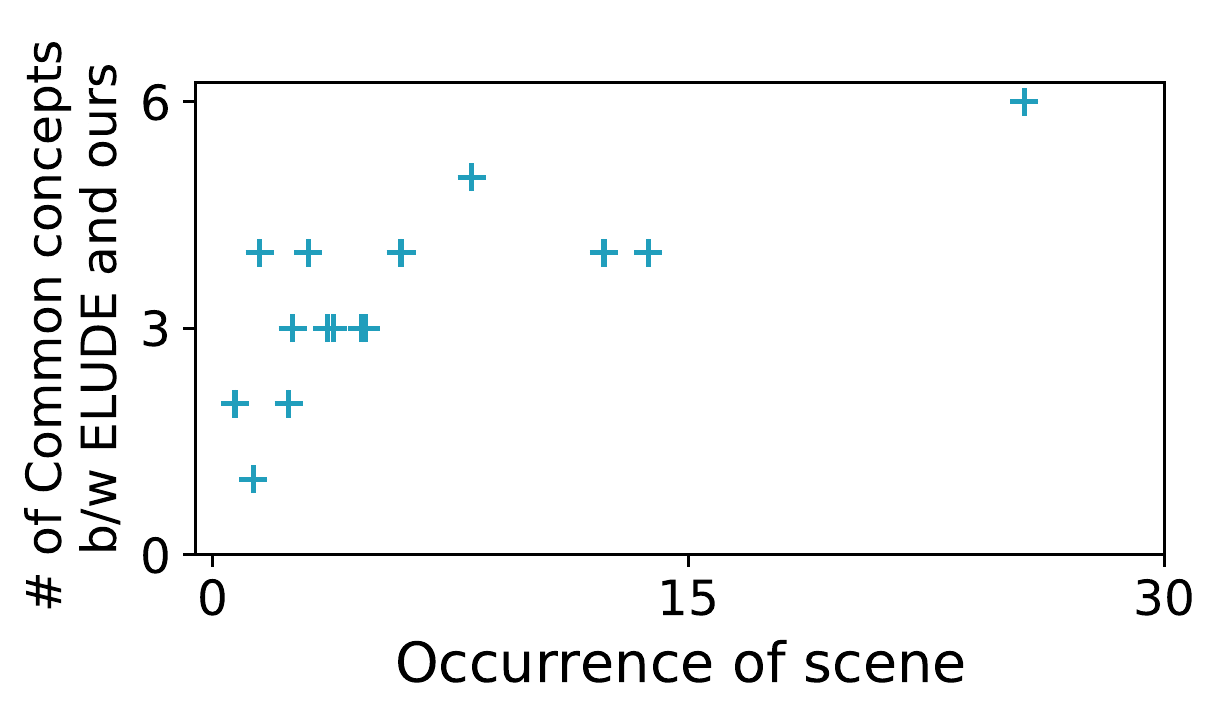}
    \caption{\textbf{Comparison with ELUDE.} The least faithful, most understandable (\MU, \LF) formulation of UFO produces explanations similar to ELUDE~\cite{ramaswamy2022elude}, particularly for scene classes that are frequent in the probe dataset. For each coarse-grained scene group, we plot the base rate of the scene class within ADE20k and the number of concepts shared between the explanations. 
    }
    \label{fig:elude_comp}
\end{figure}

Thus, we optimize~\cref{eq:lf_mu} per target class. For each coarse-grain scene group, we compare the concepts selected using this modified formulation (using $K = 8$ concepts) to the ones selected by ELUDE and report the number of common concepts between these two explanations  in~\cref{fig:elude_comp} 
For a given class, we find that the number of common concepts increases with its the base rate of the class (i.e, if sufficient positive examples of the class exist, the concepts used by ELUDE and our formulation align well (\cref{fig:elude_comp}). 

\section{Conclusion}
\label{sec:conclusion}

We present a unified method (UFO) that formalizes faithfulness and understandability as mathematical objectives and encapsulates existing concept-based explanation methods.
We show how tuning the knobs of these two objectives affects how much of the model's behavior is explained as well as how and why concept selection varies based on these objectives (e.g., more understandable explanations select concepts that are larger and more learnable ).
We also compare outputs of UFO to existing methods, and show that it is very similar to NetDissect and similar to ELUDE for well represented within the probe dataset.
Our work clearly synthesizes how existing works compare to one another for the first time and provides a useful paradigm through which future methods can be developed and described.

\smallsec{Acknowledgements} 
We thank Angelina Wang, Byron Zhang, Rohan Jinturkar and rest of the Princeton Visual AI Lab members (especially Nicole Meister) who provided helpful feedback on our work.
This material is based upon work partially supported by the National Science Foundation under Grants No. 1763642, 2145198 and 2112562. Any opinions, findings, and conclusions or recommendations expressed in this material are those of the author(s) and do not necessarily reflect the views of the National Science Foundation. We also acknowledge support from the Princeton SEAS Howard B. Wentz, Jr. Junior Faculty Award to OR, Princeton SEAS Project X Fund to RF and OR, Open Philanthropy Grant to RF, and NSF Graduate Research Fellowship to SK. 

{\small
\bibliographystyle{ieee_fullname}
\bibliography{egbib}

\begin{thebibliography}{10}\itemsep=-1pt

\bibitem{adebayo2018sanity}
Julius Adebayo, Justin Gilmer, Michael Muelly, Ian Goodfellow, Moritz Hardt,
  and Been Kim.
\newblock Sanity checks for saliency maps.
\newblock In {\em NeurIPS}, 2018.

\bibitem{adebayo2022iclr}
Julius Adebayo, Michael Muelly, Harold Abelson, and Been Kim.
\newblock Post hoc explanations may be ineffective for detecting unknown
  spurious correlation.
\newblock In {\em ICLR}, 2022.

\bibitem{adebayo2020neurips}
Julius Adebayo, Michael Muelly, Ilaria Liccardi, and Been Kim.
\newblock Debugging tests for model explanations.
\newblock In {\em NeurIPS}, 2020.

\bibitem{bau2017netdissect}
David Bau, Bolei Zhou, Aditya Khosla, Aude Oliva, and Antonio Torralba.
\newblock Network dissection: Quantifying interpretability of deep visual
  representations.
\newblock In {\em CVPR}, 2017.

\bibitem{chattopadhay2018gradcamplusplus}
Aditya Chattopadhay, Anirban Sarkar, Prantik Howlader, and Vineeth~N
  Balasubramanian.
\newblock {Grad-CAM++}: Generalized gradient-based visual explanations for deep
  convolutional networks.
\newblock In {\em WACV}, 2018.

\bibitem{dubey2022scalable}
Abhimanyu Dubey, Filip Radenovic, and Dhruv Mahajan.
\newblock Scalable interpretability via polynomials.
\newblock In {\em NeurIPS}, 2022.

\bibitem{fong19understanding}
Ruth Fong, Mandela Patrick, and Andrea Vedaldi.
\newblock Understanding deep networks via extremal perturbations and smooth
  masks.
\newblock In {\em ICCV}, 2019.

\bibitem{fong2018net2vec}
Ruth Fong and Andrea Vedaldi.
\newblock Net2vec: Quantifying and explaining how concepts are encoded by
  filters in deep neural networks.
\newblock In {\em CVPR}, 2018.

\bibitem{han2022LFA}
Tessa Han, Suraj Srinivas, and Himabindu Lakkaraju.
\newblock Which explanation should {I} choose? {A} function approximation
  perspective to characterizing post hoc explanations.
\newblock {\em arXiv:2206.01254}, 2022.

\bibitem{he2016resnet}
Kaiming He, Xiangyu Zhang, Shaoqing Ren, and Jian Sun.
\newblock Deep residual learning for image recognition.
\newblock In {\em CVPR}, 2016.

\bibitem{hoiem2012error}
Derek Hoiem, Yodsawalai Chodpathumwan, and Qieyun Dai.
\newblock Diagnosing error in object detectors.
\newblock In {\em ECCV}. Springer, 2012.

\bibitem{kim2018tcav}
Been Kim, Martin Wattenberg, Justin Gilmer, Carrie Cai, James Wexler, Fernanda
  Viegas, and Rory Sayres.
\newblock Interpretability beyond feature attribution: Quantitative testing
  with concept activation vectors ({TCAV}).
\newblock In {\em ICML}, 2018.

\bibitem{kim2022hive}
Sunnie S.~Y. Kim, Nicole Meister, Vikram~V. Ramaswamy, Ruth Fong, and Olga
  Russakovsky.
\newblock {HIVE}: Evaluating the human interpretability of visual explanations.
\newblock In {\em ECCV}, 2022.

\bibitem{kim2023helpmehelptheai}
Sunnie S.~Y. Kim, Elizabeth~Anne Watkins, Olga Russakovsky, Ruth Fong, and
  Andrés Monroy-Hernández.
\newblock {``Help me help the AI": Understanding how explainability can support
  human-AI interaction}.
\newblock In {\em CHI}, 2023.

\bibitem{KB14Adam}
Diederik~P Kingma and Jimmy Ba.
\newblock Adam: A method for stochastic optimization.
\newblock In {\em ICLR}, 2015.

\bibitem{koh2020conceptbottleneck}
Pang~Wei Koh, Thao Nguyen, Yew~Siang Tang, Stephen Mussmann, Emma Pierson, Been
  Kim, and Percy Liang.
\newblock Concept bottleneck models.
\newblock In {\em ICML}, 2020.

\bibitem{krishna2022disagreement}
Satyapriya Krishna, Tessa Han, Alex Gu, Javin Pombra, Shahin Jabbari, Steven
  Wu, and Himabindu Lakkaraju.
\newblock The disagreement problem in explainable machine learning: A
  practitioner's perspective.
\newblock {\em arXiv:2202.01602}, 2022.

\bibitem{marcos_accv_2020}
Diego Marcos, Ruth Fong, Sylvain Lobry, Rémi Flamary, Nicolas Courty, and
  Devis Tuia.
\newblock Contextual semantic interpretability.
\newblock In {\em ACCV}, 2020.

\bibitem{Petsiuk2018rise}
Vitali Petsiuk, Abir Das, and Kate Saenko.
\newblock {RISE}: Randomized input sampling for explanation of black-box
  models.
\newblock In {\em BMVC}, 2018.

\bibitem{radenovic2022neural}
Filip Radenovic, Abhimanyu Dubey, and Dhruv Mahajan.
\newblock Neural basis models for interpretability.
\newblock In {\em NeurIPS}, 2022.

\bibitem{ramaswamy2023overlookedfactors}
Vikram~V. Ramaswamy, Sunnie S.~Y. Kim, Ruth Fong, and Olga Russakovsky.
\newblock Overlooked factors in concept-based explanations: Dataset choice,
  concept salience, and human capability.
\newblock In {\em CVPR}, 2023.

\bibitem{ramaswamy2022elude}
Vikram~V. Ramaswamy, Sunnie S.~Y. Kim, Nicole Meister, Ruth Fong, and Olga
  Russakovsky.
\newblock {ELUDE}: Generating interpretable explanations via a decomposition
  into labelled and unlabelled features.
\newblock {\em arXiv:2206.07690}, 2022.

\bibitem{selvaraju2017gradcam}
Ramprasaath~R Selvaraju, Michael Cogswell, Abhishek Das, Ramakrishna Vedantam,
  Devi Parikh, and Dhruv Batra.
\newblock {Grad-CAM}: Visual explanations from deep networks via gradient-based
  localization.
\newblock In {\em ICCV}, 2017.

\bibitem{simonyan2013saliency}
Karen Simonyan, Andrea Vedaldi, and Andrew Zisserman.
\newblock Deep inside convolutional networks: Visualising image classification
  models and saliency maps.
\newblock {\em arXiv:1312.6034}, 2013.

\bibitem{sokol2020explainability}
Kacper Sokol and Peter Flach.
\newblock Explainability fact sheets: A framework for systematic assessment of
  explainable approaches.
\newblock In {\em FAccT}, 2020.

\bibitem{yuan2006model}
Ming Yuan and Yi Lin.
\newblock Model selection and estimation in regression with grouped variables.
\newblock {\em J. R. Stat. Soc}, 2006.

\bibitem{zeiler2014visualizing}
Matthew~D Zeiler and Rob Fergus.
\newblock Visualizing and understanding convolutional networks.
\newblock In {\em ECCV}, 2014.

\bibitem{zhang2016excitation}
Jianming Zhang, Sarah~Adel Bargal, Zhe Lin, Jonathan Brandt, Xiaohui Shen, and
  Stan Sclaroff.
\newblock Top-down neural attention by excitation backprop.
\newblock {\em IJCV}, 2018.

\bibitem{zhou2016cam}
Bolei Zhou, Aditya Khosla, Agata Lapedriza, Aude Oliva, and Antonio Torralba.
\newblock Learning deep features for discriminative localization.
\newblock In {\em CVPR}, 2016.

\bibitem{zhou2017places}
Bolei Zhou, Agata Lapedriza, Aditya Khosla, Aude Oliva, and Antonio Torralba.
\newblock Places: A 10 million image database for scene recognition.
\newblock {\em TPAMI}, 40, 2017.

\bibitem{zhou2018ibd}
Bolei Zhou, Yiyou Sun, David Bau, and Antonio Torralba.
\newblock Interpretable basis decomposition for visual explanation.
\newblock In {\em ECCV}, 2018.

\bibitem{zhou2017ade20k}
Bolei Zhou, Hang Zhao, Xavier Puig, Sanja Fidler, Adela Barriuso, and Antonio
  Torralba.
\newblock Scene parsing through {ADE20k} dataset.
\newblock In {\em CVPR}, 2017.

\bibitem{zhou2019ade20k_ijcv}
Bolei Zhou, Hang Zhao, Xavier Puig, Tete Xiao, Sanja Fidler, Adela Barriuso,
  and Antonio Torralba.
\newblock Semantic understanding of scenes through the {ADE20k} dataset.
\newblock {\em IJCV}, 2019.

\end{thebibliography}
}

\appendix

In this appendix, we provide more details about our method, as well as some additional results. 

\section{Additional details about comparisons with prior work}
\label{sec:netdissect}
Here we provide more details about the different methods, and how they can be thought of within our framework. 

\smallsec{Net2Vec and TCAV}
For Net2Vec~\cite{fong2018net2vec} and TCAV~\cite{kim2018tcav}, the authors align the feature space with concepts, without considering the final output. This can be achieved with $\lambda_1 = 0$.
The authors allow the features $f$ to be any layer within the trained model $M$, and learn $h_\text{conc}$ as a series of individual indicator functions to the concepts for each neuron in $f$ or as a linear combination of the different neurons in $f$, i.e, $h_\text{conc} \colon \mathbb{R}^n \rightarrow \mathbb{R}^C$, where $n$ is the number of neurons in a layer of the CNN and $C$ is the number of concepts. The authors allow the selection function $S$ to select all the $C$ concepts, i.e, $K=C$. Thus, these works optimize $L_\text{align}$:
\begin{align}
    \label{eq:net2vec_tcav}
    L_\text{align}^{\text{Net2Vec}, \text{TCAV}} & = \sum_{j \in \{1, 2, \ldots C\}}\sum_{x\in X} CE(A(x)_j, (h_\text{conc}\circ f(x))_j)
\end{align}
Net2Vec also considers aligning individual neurons with concepts. This can be achieved by forcing $h_\text{conc}$ to be an indicator function: each neuron is aligned with exactly one concept.  

\smallsec{NetDissect}
NetDissect~\cite{bau2017netdissect} uses a slightly different framework compared to ours, however, we show that by rewriting $L_\text{align}$, we can consider NetDissect within our framework. We first rewrite NetDissect using the following notation. 
\begin{itemize}
    \item Suppose $A_{\text{seg}} \colon X \to \mathbb{R}^{C \times H\times W}$ is the segmentation map for $C$ concepts.
    \item $f \colon \mathcal{X} \to \mathbb{R}^{n \times H' \times W'}$ is the feature space. 
    \item $t: \mathbb{R}^{n \times H' \times W'} \rightarrow \{0,1\}^{D \times H \times W}$.
    This is an upsampling and thresholding function: first, the vector is bilinearly upsampled to size $H \times W$ for each neuron, and thresholded such that only the top $0.5\%$ for a neuron is activated.
\end{itemize}
Now, for each neuron $i \in \{1, 2, \ldots n\}$, they compute the concept $j$ that maximizes
\begin{align}
\label{eq:netdissect}
\text{IOU}_i 
   := & \argmax_{j \in \{1, 2, \ldots C\}} \left(\frac{\sum_{x \in X} ((A_{\text{seg}}(x))_j \cap (t \circ f(x))_i)}{\sum_{x \in X} ((A_{\text{seg}}(x))_j \cup (t \circ f(x))_i)}\right)
\end{align}

In order to consider NetDissect within our framework, we can rewrite $L_\text{align}$ as follows. We first consider $f_i$ at a single neuron $i$, i.e. $f_i \colon \mathcal{X} \rightarrow \mathbb{R}^{H' \times W'}$. Then, rather than using $\sum_x CE(A(x), h_\text{conc}\circ f(x))$, we can express $L_\text{align}$ in terms of \cref{eq:netdissect}, with $|S| = 1$:

\begin{align}
    L_{align} = - \sum_{j \in \{1, 2, \ldots C\}} \mathbbm{1}_S \circ \text{IOU}_i
\end{align}

\smallsec{IBD}
For IBD~\cite{zhou2018ibd}, $h_\text{conc}$ is a linear combination of the activations of each neuron and $h_\text{pred}$ is a linear combination of the outputs of $h_\text{conc}$, very similar to our \textbf{sf,su} framework. The main difference is in an additional constraint imposed on $h_\text{pred}$: that the coefficients are all non-negative, and each target class is allowed to use exactly $K$ concepts (but these do not need to be the same across target classes). Thus, for IBD, $L_\text{mimic}$ can be written as:

\begin{align}
    \label{eq:ibd}
    \forall i \in \{1, 2, \ldots, D\} & \nonumber\\
 \left(L_\text{mimic}^\text{IBD}\right)_i & = 
   \sum_{x \in X} \|(g \circ f(x))_i -  \nonumber \\ & h_\text{pred}^i  \circ \mathbbm{1}_{S_i} \circ 
   p \circ h_\text{conc}\circ f(x)
   \|
   \\
   \text{such that } \nonumber\\
&   h_\text{pred}^i(x) = W_i^Tf(x) \nonumber \\
 & \textcolor{red}{ W_{i, k}\geq 0} \ \ \ \forall k \in \{1, 2, \ldots n\} \nonumber
\end{align}

As mentioned in the main text Section 5.2, the non-negative constraint changes the concepts chosen: examples of concepts chosen are in table~\ref{tab:ibd}

\begin{table*}[t]
\centering
\begin{tabular}{L{2cm} L{6.5cm}L{6.5cm}}
\toprule
scene & IBD & UFO(Ours) \\
\toprule
attic & heater, basket, stairway, breads, magazine, television camera, drum & backpack, grandstand, coffee maker, pitcher, microwave, sand, door, sculpture\\ \arrayrulecolor{black!30}\midrule
bathroom & screen door, village, water tower, tray, candelabrum, stands, drinking glass & grandstand, backpack, crt screen, bench, microwave, double door, sculpture, work surface\\\midrule
bedchamber & headboard, pillow, shade, vault, eiderdown, water tower, tent & coffee maker, pitcher, spotlight, microwave, cabinet, door, sky, sculpture\\\midrule
bedroom & headboard, pillow, eiderdown, shade, lower sash, shower, shirt & grandstand, coffee maker, doorframe, ladder, sculpture, spotlight, work surface, clock\\\midrule
conference-room & bulletin board, wineglass, trouser, escalator, mouse, button panel, mouse pad & grandstand, platform, counter, pitcher, microwave, ladder, floor, desk\\\midrule
crosswalk & vineyard, traffic light, autobus, chain wheel, trailer, skylight, cockpit & coffee maker, grandstand, land, platform, crt screen, backpack, doorframe, television\\\midrule
dining-room & chandelier, candelabrum, back pillow, skirt, carpet, cart, grand piano & coffee maker, pitcher, platform, chest, windowpane, road, door, table\\\midrule
downtown & skyscraper, paper towel, gas station, candelabrum, place mat, slot machine, crosswalk & grandstand, land, pitcher, platform, crt screen, television, sky, sofa\\\midrule
highway & catwalk, autobus, document, book stand, dashboard, slats, corner pocket & coffee maker, platform, plant, flag, sky, lamp, door, table\\\midrule
hotel-room & bed, tracks, candlestick, cushion, seat cushion, capital, candle & land, microwave, spotlight, cabinet, sculpture, dishwasher, clock, work surface\\\midrule
kitchen & stove, refrigerator, tray, kitchen island, container, screen door, microwave & grandstand, coffee maker, backpack, crt screen, pitcher, doorframe, cap, faucet\\\midrule
living-room & post, cushion, riser, fireplace, monitoring device, sconce, bumper & pitcher, doorframe, counter, spotlight, windowpane, cabinet, door, table\\\midrule
parking-garage/outdoor & paper towel, crane, windows, notebook, steam shovel, gym shoe, television & coffee maker, grandstand, land, backpack, platform, crt screen, doorframe, television\\\midrule
recreation-room & pool table, court, microwave, table football, slot machine, wire, grand piano & grandstand, land, chest, counter, spotlight, floor, windowpane, sky\\\midrule
residential-neighborhood & sill, balloon, trailer, metal shutters, flowerpot, switch, synthesizer & coffee maker, faucet, land, platform, doorframe, sculpture, dishwasher, floor\\\midrule
skyscraper & skyscraper, display board, workbench, manhole, paw, lighthouse, gas station & coffee maker, land, pitcher, television, sky, sofa, lamp, spotlight\\\midrule
street & slats, roundabout, crosswalk, beak, arcades, bus, parking & coffee maker, land, faucet, platform, crt screen, microwave, doorframe, television\\\midrule
television-room & seat base, brick, sash, inside arm, gravel, water wheel, pantry & pitcher, chest, cap, microwave, spotlight, counter, desk, door\\\midrule
waiting-room & armchair, sconce, shoe, console table, back pillow, canvas, dishrag & pitcher, chest, counter, spotlight, sky, doorframe, sofa, table\\\midrule
youth-hostel & sweater, towel, equipment, kettle, wardrobe, vent, partition & grandstand, doorframe, ladder, microwave, spotlight, sculpture, work surface, flag\\ \arrayrulecolor{black}\bottomrule
\end{tabular}
\caption{\textbf{Concepts chosen by IBD~\cite{zhou2018ibd} versus that chosen by our method.} We see that the non-negative constraint added by IBD changes the concepts chosen by quite a lot.}
\label{tab:ibd}
\end{table*}


\section{More results}
\label{sec:more_res}
In this section, we give more details about our experiment set up and highlight additional results. 

\smallsec{Experimental setup}
We use a greedy method to select uncorrelated concepts, using Pearson's correlation coefficient.  For each understandability setting, we compute the correlation coefficient between all pairs of concepts (using either the base rates or the learned concept vector). Next, we compute the 90\% percentile correlation coefficient and set that as a threshold. We add each concept to the list of chosen concepts if it is not more correlated than the threshold with any of the previously chosen concepts. 

When choosing the values of $\lambda_1$ and $\lambda_2$, we fix $\lambda_1$ to 1, and pick $\lambda_2$ from $\{1.0, 0.5, 0.1, 0.05, 0.01, 0.005\}$ that best explains the model on the validation set. 

\smallsec{Concepts for coarse grained scenes}
We first report the concepts chosen across the 6 faithfulness-understandability settings described in the main text. Similar to the fine-grained model, we see that the attributes chosen vary based on the setting (Tab.~\ref{tab:grouped_attr_comp}).

\begin{table*}[t]
    \centering
    \resizebox{0.82\linewidth}{!}{
    \begin{tabular}{L{2cm} l   L{4cm}  L{4cm}  L{4cm}}
 \toprule
& & {most understandable (\MU)} & {somewhat understandable (\SU)} & {least understandable (\LU)} \\ 
\midrule 
shopping-dining & \SF & \textit{\textcolor{red}{bed}}, \textcolor{NavyBlue}{bulletin board}, \textit{\textcolor{red}{sky}} & \textcolor{NavyBlue}{\textbf{cap}}, \textit{\textcolor{red}{footbridge}}, \textcolor{NavyBlue}{baby buggy} & \textit{\textcolor{red}{\textbf{faucet}}}, \textit{\textcolor{red}{land}}, \textcolor{NavyBlue}{\textbf{cap}}\\
 & \LF & \textit{\textcolor{red}{grandstand}}, \textit{\textcolor{red}{pillow}}, \textit{\textcolor{red}{bird}} & \textit{\textcolor{red}{table tennis}}, \textcolor{NavyBlue}{\textbf{cap}}, \textit{\textcolor{red}{loudspeaker}} & \textit{\textcolor{red}{\textbf{faucet}}}, \textcolor{NavyBlue}{pitcher}, \textcolor{NavyBlue}{counter}\\
\midrule
workplace & \SF & \textit{\textcolor{red}{\textbf{sky}}}, \textcolor{NavyBlue}{floor}, \textcolor{NavyBlue}{desk} & \textcolor{NavyBlue}{\textbf{dog}}, \textcolor{NavyBlue}{footbridge}, \textit{\textcolor{red}{\textbf{baby buggy}}} & \textcolor{NavyBlue}{platform}, \textit{\textcolor{red}{bread}}, \textit{\textcolor{red}{jar}}\\
 & \LF & \textit{\textcolor{red}{\textbf{sky}}}, \textit{\textcolor{red}{bed}}, \textit{\textcolor{red}{tree}} & \textit{\textcolor{red}{\textbf{baby buggy}}}, \textcolor{NavyBlue}{\textbf{dog}}, \textit{\textcolor{red}{lake}} & \textcolor{NavyBlue}{paper}, \textcolor{NavyBlue}{bulletin board}, \textcolor{NavyBlue}{runway}\\
\midrule
home-hotel & \SF & \textcolor{NavyBlue}{bed}, \textcolor{NavyBlue}{towel}, \textcolor{NavyBlue}{floor} & \textit{\textcolor{red}{\textbf{footbridge}}}, \textit{\textcolor{red}{cap}}, \textit{\textcolor{red}{railroad train}} & \textcolor{NavyBlue}{table tennis}, \textcolor{NavyBlue}{\textbf{blind}}, \textcolor{NavyBlue}{bread}\\
 & \LF & \textit{\textcolor{red}{grandstand}}, \textit{\textcolor{red}{road}}, \textit{\textcolor{red}{bird}} & \textit{\textcolor{red}{bulletin board}}, \textit{\textcolor{red}{text}}, \textit{\textcolor{red}{\textbf{footbridge}}} & \textcolor{NavyBlue}{\textbf{blind}}, \textcolor{NavyBlue}{faucet}, \textit{\textcolor{red}{paper}}\\
\midrule
indoor-transport & \SF & \textcolor{NavyBlue}{floor}, \textit{\textcolor{red}{tree}}, \textcolor{NavyBlue}{work surface} & \textit{\textcolor{red}{\textbf{table tennis}}}, \textcolor{NavyBlue}{\textbf{railroad train}}, \textit{\textcolor{red}{\textbf{cap}}} & \textcolor{NavyBlue}{\textbf{clock}}, \textit{\textcolor{red}{\textbf{cap}}}, \textcolor{NavyBlue}{\textbf{railroad train}}\\
 & \LF & \textit{\textcolor{red}{grandstand}}, \textit{\textcolor{red}{dog}}, \textit{\textcolor{red}{umbrella}} & \textit{\textcolor{red}{\textbf{table tennis}}}, \textit{\textcolor{red}{\textbf{cap}}}, \textit{\textcolor{red}{lake}} & \textcolor{NavyBlue}{backpack}, \textcolor{NavyBlue}{\textbf{clock}}, \textcolor{NavyBlue}{runway}\\
\midrule
indoor-cultural & \SF & \textit{\textcolor{red}{sky}}, \textit{\textcolor{red}{\textbf{work surface}}}, \textcolor{NavyBlue}{desk} & \textcolor{NavyBlue}{lake}, \textit{\textcolor{red}{footbridge}}, \textcolor{NavyBlue}{dog} & \textcolor{NavyBlue}{sculpture}, \textit{\textcolor{red}{trunk}}, \textit{\textcolor{red}{\textbf{runway}}}\\
 & \LF & \textit{\textcolor{red}{\textbf{work surface}}}, \textit{\textcolor{red}{towel}}, \textcolor{NavyBlue}{central reservation} & \textit{\textcolor{red}{land}}, \textit{\textcolor{red}{microwave}}, \textit{\textcolor{red}{exhaust hood}} & \textcolor{NavyBlue}{paper}, \textit{\textcolor{red}{\textbf{runway}}}, \textit{\textcolor{red}{television}}\\
\midrule
water, ice, snow & \SF & \textcolor{NavyBlue}{\textbf{lake}}, \textcolor{NavyBlue}{mountain}, \textit{\textcolor{red}{floor}} & \textcolor{NavyBlue}{\textbf{lake}}, \textcolor{NavyBlue}{footbridge}, \textcolor{NavyBlue}{cap} & \textcolor{NavyBlue}{\textbf{land}}, \textcolor{NavyBlue}{backpack}, \textcolor{NavyBlue}{refrigerator}\\
 & \LF & \textit{\textcolor{red}{grandstand}}, \textit{\textcolor{red}{door}}, \textit{\textcolor{red}{scaffolding}} & \textit{\textcolor{red}{\textbf{horse}}}, \textcolor{NavyBlue}{\textbf{lake}}, \textit{\textcolor{red}{text}} & \textcolor{NavyBlue}{counter}, \textcolor{NavyBlue}{\textbf{land}}, \textit{\textcolor{red}{\textbf{horse}}}\\
\midrule
mountains, hills, desert & \SF & \textcolor{NavyBlue}{\textbf{mountain}}, \textcolor{NavyBlue}{sky}, \textit{\textcolor{red}{building}} & \textcolor{NavyBlue}{lake}, \textcolor{NavyBlue}{cap}, \textcolor{NavyBlue}{\textbf{land}} & \textcolor{NavyBlue}{backpack}, \textcolor{NavyBlue}{\textbf{land}}, \textcolor{NavyBlue}{\textbf{mountain}}\\
 & \LF & \textit{\textcolor{red}{bird}}, \textit{\textcolor{red}{windowpane}}, \textit{\textcolor{red}{grandstand}} & \textit{\textcolor{red}{footbridge}}, \textit{\textcolor{red}{text}}, \textit{\textcolor{red}{signboard}} & \textit{\textcolor{red}{flag}}, \textcolor{NavyBlue}{\textbf{land}}, \textcolor{NavyBlue}{\textbf{mountain}}\\
\midrule
forest, field, jungle & \SF & \textcolor{NavyBlue}{tree}, \textit{\textcolor{red}{building}}, \textit{\textcolor{red}{road}} & \textit{\textcolor{red}{footbridge}}, \textcolor{NavyBlue}{\textbf{trunk}}, \textcolor{NavyBlue}{\textbf{horse}} & \textcolor{NavyBlue}{\textbf{horse}}, \textcolor{NavyBlue}{\textbf{trunk}}, \textit{\textcolor{red}{ladder}}\\
 & \LF & \textit{\textcolor{red}{central reservation}}, \textit{\textcolor{red}{\textbf{bulletin board}}}, \textit{\textcolor{red}{\textbf{spotlight}}} & \textcolor{NavyBlue}{\textbf{trunk}}, \textcolor{NavyBlue}{\textbf{horse}}, \textit{\textcolor{red}{\textbf{spotlight}}} & \textit{\textcolor{red}{\textbf{bulletin board}}}, \textcolor{NavyBlue}{\textbf{horse}}, \textit{\textcolor{red}{forecourt}}\\
\midrule
outdoor-transport & \SF & \textcolor{NavyBlue}{road}, \textcolor{NavyBlue}{sky}, \textcolor{NavyBlue}{building} & \textit{\textcolor{red}{\textbf{lake}}}, \textit{\textcolor{red}{\textbf{cap}}}, \textcolor{NavyBlue}{footbridge} & \textcolor{NavyBlue}{\textbf{runway}}, \textcolor{NavyBlue}{bulletin board}, \textcolor{NavyBlue}{ladder}\\
 & \LF & \textit{\textcolor{red}{desk}}, \textit{\textcolor{red}{rack}}, \textit{\textcolor{red}{towel}} & \textit{\textcolor{red}{\textbf{lake}}}, \textit{\textcolor{red}{\textbf{cap}}}, \textit{\textcolor{red}{\textbf{horse}}} & \textcolor{NavyBlue}{\textbf{runway}}, \textcolor{NavyBlue}{pitcher}, \textit{\textcolor{red}{\textbf{horse}}}\\
\midrule
cultural-historic & \SF & \textcolor{NavyBlue}{building}, \textit{\textcolor{red}{\textbf{lake}}}, \textcolor{NavyBlue}{sky} & \textcolor{NavyBlue}{baby buggy}, \textit{\textcolor{red}{\textbf{lake}}}, \textcolor{NavyBlue}{\textbf{trunk}} & \textcolor{NavyBlue}{forecourt}, \textit{\textcolor{red}{\textbf{trunk}}}, \textit{\textcolor{red}{table tennis}}\\
 & \LF & \textit{\textcolor{red}{fluorescent}}, \textit{\textcolor{red}{cabinet}}, \textit{\textcolor{red}{bed}} & \textit{\textcolor{red}{\textbf{lake}}}, \textit{\textcolor{red}{horse}}, \textit{\textcolor{red}{railroad train}} & \textit{\textcolor{red}{pitcher}}, \textit{\textcolor{red}{blind}}, \textit{\textcolor{red}{runway}}\\
\midrule
sports fields, parks & \SF & \textcolor{NavyBlue}{grandstand}, \textit{\textcolor{red}{desk}}, \textcolor{NavyBlue}{tree} & \textcolor{NavyBlue}{\textbf{baby buggy}}, \textit{\textcolor{red}{lake}}, \textit{\textcolor{red}{cap}} & \textcolor{NavyBlue}{\textbf{baby buggy}}, \textcolor{NavyBlue}{\textbf{net}}, \textit{\textcolor{red}{paper}}\\
 & \LF & \textit{\textcolor{red}{scaffolding}}, \textit{\textcolor{red}{air conditioner}}, \textit{\textcolor{red}{curtain}} & \textcolor{NavyBlue}{\textbf{baby buggy}}, \textit{\textcolor{red}{telephone booth}}, \textit{\textcolor{red}{lake}} & \textcolor{NavyBlue}{\textbf{net}}, \textcolor{NavyBlue}{table}, \textit{\textcolor{red}{double door}}\\
\midrule
cabins, gardens, farms & \SF & \textcolor{NavyBlue}{tree}, \textcolor{NavyBlue}{plant}, \textit{\textcolor{red}{bulletin board}} & \textit{\textcolor{red}{\textbf{table tennis}}}, \textit{\textcolor{red}{cap}}, \textcolor{NavyBlue}{dog} & \textit{\textcolor{red}{\textbf{table tennis}}}, \textcolor{NavyBlue}{blind}, \textit{\textcolor{red}{backpack}}\\
 & \LF & \textit{\textcolor{red}{bird}}, \textit{\textcolor{red}{spotlight}}, \textit{\textcolor{red}{scaffolding}} & \textit{\textcolor{red}{land}}, \textit{\textcolor{red}{lake}}, \textit{\textcolor{red}{railroad train}} & \textcolor{NavyBlue}{bread}, \textit{\textcolor{red}{counter}}, \textit{\textcolor{red}{poster}}\\
\midrule
comm-buildings/towns & \SF & \textcolor{NavyBlue}{building}, \textit{\textcolor{red}{\textbf{grandstand}}}, \textcolor{NavyBlue}{road} & \textcolor{NavyBlue}{telephone booth}, \textcolor{NavyBlue}{trunk}, \textit{\textcolor{red}{\textbf{land}}} & \textcolor{NavyBlue}{\textbf{forecourt}}, \textcolor{NavyBlue}{platform}, \textit{\textcolor{red}{net}}\\
 & \LF & \textit{\textcolor{red}{\textbf{grandstand}}}, \textit{\textcolor{red}{desk}}, \textit{\textcolor{red}{piano}} & \textit{\textcolor{red}{\textbf{land}}}, \textit{\textcolor{red}{horse}}, \textit{\textcolor{red}{lake}}& \textcolor{NavyBlue}{\textbf{forecourt}}, \textcolor{NavyBlue}{poster}, \textcolor{NavyBlue}{flag}\\
 \bottomrule

\end{tabular}}
    \caption{\textbf{Selected concepts vary based on faithfulness-understandability setting.} Similar to the main text Tab. 2., we examine the concepts chosen for each scene group across 6 settings (\{\MU, \SU, \LU\} $\times$ \{\SF, \LF\}. We report the 3 concepts with highest absolute weights within the explanation. Common concepts are \textbf{bolded}, \textit{\textcolor{red}{Red}} denotes that the coefficient is negative, whereas \textcolor{NavyBlue}{blue} denotes that the coefficient is positive. We note that the concepts highlighted are typically not shared among different explanations  }
    \label{tab:grouped_attr_comp}
\end{table*}

\end{document}